\documentclass[12pt]{colt2018}

\coltauthor{\Name{Zac Cranko}\\
    \addr The Australian National University \& Data 61
    \AND
    \Name{Simon Kornblith}\\
    \addr Google Brain
    \AND
    \Name{Zhan Shi}\\
    \addr The University of Illinois at Chicago
    \AND
    \Name{Richard Nock}\\
    \addr Data 61, The Australian National University \& The University of Sydney  
}
\makeatletter
\let\@jmlr@check@packages\relax
\makeatother

\usepackage{./macros}

\usepackage{times}
\usepackage{calc}

\usepackage{xspace}
\usepackage{url}
\usepackage[]{algorithm2e}

\definecolor{red}    {HTML}{c82829} 
\definecolor{yellow} {HTML}{eab700} 
\definecolor{green}  {HTML}{718c00} 
\definecolor{cyan}   {HTML}{3e999f} 
\definecolor{blue}   {HTML}{4271ae} 
\definecolor{magenta}{HTML}{8959a8} 

\usepackage{tikz}
\usetikzlibrary{arrows, calc, through, positioning}
\tikzset{
    dot/.style={draw,circle,minimum size=3pt,inner sep=0pt,outer sep=0pt,fill=black}
}

\title{Lipschitz Networks and Distributional Robustness}

\ExplSyntaxOn

\makeatletter
\NewDocumentCommand{\multicitep}{m}
 {
  \NAT@open
  \mjb_multicitep:n { #1 }
  \NAT@close
 }
\makeatother

\seq_new:N \l_mjb_multicite_in_seq
\seq_new:N \l_mjb_multicite_out_seq
\seq_new:N \l_mjb_cite_seq

\cs_new_protected:Npn \mjb_multicitep:n #1
 {
  \seq_set_split:Nnn \l_mjb_multicite_in_seq { ; } { #1 }
  \seq_clear:N \l_mjb_multicite_out_seq
  \seq_map_inline:Nn \l_mjb_multicite_in_seq
   {
    \mjb_cite_process:n { ##1 }
   }
  \seq_use:Nn \l_mjb_multicite_out_seq { ;~ }
 }

\cs_new_protected:Npn \mjb_cite_process:n #1
 {
  \seq_set_split:Nnn \l_mjb_cite_seq { , } { #1 }
  \int_compare:nTF { \seq_count:N \l_mjb_cite_seq == 1 }
   {
    \seq_put_right:Nn \l_mjb_multicite_out_seq
     { \citeauthor{#1},~\citeyear{#1} }
   }
   {
    \seq_put_right:Nx \l_mjb_multicite_out_seq
     {
      \exp_not:N \citeauthor{\seq_item:Nn \l_mjb_cite_seq { 1 }},~
      \exp_not:N \citeyear{\seq_item:Nn \l_mjb_cite_seq { 1 }},~
      \seq_item:Nn \l_mjb_cite_seq { 2 }
     }
   }
 }
\ExplSyntaxOff

\def\probm{\mathscr P}

\def\cost{\mathcaleu C}
\def\costb{\mathcaleu B}

\def\risk{\operatorname{risk}_\ell}

\def\neuralnet{\upPhi}

\let\coup\upPi
\def\dirac{\delta}
\def\bmeas{\mathrm M}

\def\ellf{\bgroup\ell_f\egroup}

\def\opnorm#1{\lnorm{#1}_{\mathrm{op}}}
\let\lnorm\norm
\let\norm\abs

\newlist{paperlayout}{itemize*}{1}
\setlist*[paperlayout,1]{%
  label={},
  before={},
  itemjoin={;\; }, 
  itemjoin*={;\; and finally, },
  after={.}
}

\newlist{letblock}{itemize*}{1}
\setlist*[letblock,1]{%
  label={},
  before={Let },
  itemjoin={;\allowbreak }, 
  itemjoin*={; \allowbreak{}and },
  after={.}
}

\begin{document}

\maketitle

\begin{abstract}
    Robust risk minimisation has several advantages: it has been studied with regards to improving the generalisation properties of models and robustness to adversarial perturbation. We bound the distributionally robust risk for a model class rich enough to include deep neural networks by a regularised empirical risk involving the Lipschitz constant of the model. This allows us to interpret and quantify the robustness properties of a deep neural network. As an application we show the distributionally robust risk upperbounds the adversarial training risk.
\end{abstract}

\section{Introduction}

Classical risk minimisation picks a model $f$ to minimise risk with respect to a loss function $\ell$ and distribution $\mu$. It's convenient to combine the loss function and model into a \emph{parameterised loss function} $\ellf$ where the exact nature of the parameterisation is left purposefully ambiguous. The \emph{risk} of the model $f$ with respect to $\mu$ is
\begin{gather}
   \risk(f;\mu) \defas \int \mu(\d s)\ellf(s).
\end{gather}
The \emph{distributionally robust risk}, however, is defined for a collection of distributions, $\cal P$, called the \emph{uncertainty set} \citep{duchi2016statistics,sinha2017certifiable,shafieezadeh2017regularization,blanchet2016robust,blanchet2017data} 
\begin{gather}
    \risk(f;\cal P) \defas \sup_{\nu\in\cal P} \int \nu(\d s)\ellf(s).\label{defn:distributionally_robust_risk}
\end{gather}

When the uncertainty set is a transportation cost- or Wasserstein-ball containing $\mu$ (defined formally in \autoref{sec:preliminaries}), under certain technical conditions, \eqref{defn:distributionally_robust_risk} has the dual form \citep{blanchet2016robust,blanchet2017data,gao2016distributionally,shafieezadeh2017regularization} 
\begin{gather}
    \risk(f;\cal P) = \risk(f;\mu) + L\rbr{f}. \label{eq:reguarised_risk}
\end{gather}
involving only the empirical risk and a regularisation term that reflects the uncertainty set. However, the conditions required of $\ell$ and model class to obtain a representation of the form \eqref{eq:reguarised_risk} are typically onerous. Alternative approaches to robust risk minimisation instead upperbound \eqref{defn:distributionally_robust_risk}, however attempts to do this  typically result in upper bounds which are not closed form  \citep{sinha2017certifiable,raghunathan2018certified} and thus difficult to minimise. For a broad class of models $f$ (including deep models) and neighbourhoods $\cal P$, we upperbound \eqref{defn:distributionally_robust_risk} with an objective function of the form \eqref{eq:reguarised_risk} wherein $L$ is a function depending on $\ell$ and $\cal P$ that can be related to the Lipschitz constant of the model $f$. This formally justifies the some of the success of Lipschitz regularised or constrained neural networks \citep{yoshida2017spectral,cisse2017parseval,miyato2018spectral,gouk2018regularisation} and completely specifies the otherwise heuristically chosen regularisation parameter.

The rest of the paper is laid out as follows: 
\begin{paperlayout}[before={}]
    \item in \autoref{sec:preliminaries} we define notation and give some definitions 
    \item in \autoref{sec:robust_linear_classification} we generalise the binary classification result of \citet[Theorem~3.11]{shafieezadeh2017regularization} to an arbitrary metric space of labels (\autoref{thm:multiclass_robust_dual_minimisation}) that is, a space general enough to be sufficient for multiclass problems
    \item in \autoref{sec:robust_deep_classification} we upperbound the deep robust classification problem by a convex problem (\autoref{thm:pushforward_risk}) which we call the \emph{pushforward risk minimisation problem}, to which we apply \autoref{thm:multiclass_robust_dual_minimisation} to compute the (closed form) dual formulation (\autoref{prop:deep_robust_classifcation})
    \item in \autoref{ssec:distributional_robustness_is_adversarial_robustness} we observe that the adversarial saddle-point objective of \citet{madry2017towards} is upper bounded by the distributionally robust risk (\autoref{thm:adversarial_risk_bound}) 
    \item we finish with a discussion of other recent results in this area \autoref{sec:discussion} and conclusion \autoref{sec:conclusion}
\end{paperlayout}

\section{Preliminaries}\label{sec:preliminaries}

The extended real line is $\Rx \defas [-\infty, +\infty]$, the Iverson bracket $\iver{P}$ returns 1 if a proposition $P$ is true and $0$ otherwise. For a natural number $n$, define the set $[n]\defas\cbr{1,\dots,n}$.
In what follows $S$, $T$ are topological spaces, with continuous duals $S^*$, $T^*$. The collection of Borel probability measures on $S$ is $\probm(S)$. The set $\bmeas(S;T)$ is the collection of Borel mappings $S\to T$. 
The Fenchel conjugate of a function $f:S\to\Rx$ is $f^*:S^*\to\Rx$ with $f^*(s^*) \defas \sup_{s\in S}\rbr{\inp{s^*,s} - f(s)}$.
For a mapping $f\in\bmeas(S;T)$ and a measure $\mu\in\probm(S)$ the \emph{push forward of $\mu$ by $f$} is the measure $f_\#\mu \in\probm(T)$ with $f_\#\mu \defas \mu \circ f^{-1}$. 
The Dirac measure at a point $s \in S$ is $\dirac_s(A) \defas \iver{s\in A}$. Fix $\mu\in\probm(S)$ and suppose we have sample $(s_i)_{i=1}^n\sim\mu$; a \emph{$n$-realisation of $\mu$}, $\hat\mu(n)\defas\frac1n\sum_{i=1}^n\dirac_{s_i} \in\probm(S)$, is an $n$-size empirical distribution (itself a random variable which we consider fixed) corresponding to $\mu$.
%
The collection of \emph{$(\mu,\nu)$-couplings} is
\begin{gather}
    \coup(\mu,\nu) \defas \cbr{
        \pi\in\probm(S\times T)
        :
        \mu = \int_{T}\pi(\marg, \d t),\:\nu = \int_{S}\pi(\d s, \marg )
    }.
\end{gather}

%

%

%


\begin{letblock}
    \item $c\in\bmeas(S\times S; \Rx)$ 
\end{letblock}
The \emph{$c$-transport cost} of $\mu,\nu\in\probm(S)$, and \emph{$c$-transport cost ball} of radius $\rho\geq0$ centred at $\mu\in\probm(S)$ are respectively 
\begin{gather}
    \cost_{c}(\mu,\nu) \defas \inf_{\pi\in\coup(\mu,\nu)} \int_{S\times S} c \d\pi,\and
    \costb_{c, \rho}(\mu) \defas \cbr{\nu \in \probm(S) \given \cost_c(\mu,\nu) ≤  \rho}.
\end{gather}
When $d$ is a metric on $S$, $\cost_{d}(\mu,\nu)$ is called a \emph{Wasserstein distance}, and $\costb_{d, \rho}(\mu) $ is called a \emph{Wasserstein ball}. A function $f$ between metric spaces $(S, d_S)$ and $(T, d_T)$ is \emph{Lipschitz} if there exists a real number $L\geq 0$ such that
\begin{gather}
    \forall{s,s'\in S} d_T(f(s), f(s')) ≤ L d_S(s, s').\inline{Let }\lip(f) \defas \sup_{s≠s'}\frac{d_T(f(s), f(s'))}{d_S(s, s')}.
\end{gather}
If $S$ admits a decomposition $S = X\times Y$ where $X$ and $Y$ are metric spaces themselves. Then $\lips_X(f) \defas \sup_{y\in Y} \lip(f(\marg, y))$.

A \emph{parameterised loss function} (by a model $f\in\cal F$) is a mapping $\ellf: S\to\Rx$, examples of which include
\begin{itemize}
    \item Classification. \begin{letblock}[before={Define }]
        \item the input--label space $S \defas X\times Y$ with $Y$ discrete
        \item $\cal F$ family of functions $X\to \probm(Y)$
        \item $\ell:\probm(Y)\times Y\to\Rx$ is a scoring rule
    \end{letblock}
    Then $\ellf(x,y) = \ell(f(x), y)$.
    \item Regression. \begin{letblock}[before={Define }]
        \item the input--response space $S \defas X\times Y$
        \item $\cal F$ family of functions $X\to Y$
        \item $\ell\defas d$, some metric on $Y$
    \end{letblock}
    Then $\ellf(x,y) = d(f(x), y)$.
    \item Density estimation. \begin{letblock}[before={Define }]
        \item $S\defas\R^n$
        \item $\cal F$ family of functions $S\to \R_+$
        \item $\ell\defas -\log$
    \end{letblock}
    Then $\ellf(x) = -\log f(x)$.
\end{itemize}




\section{Robust linear classification}\label{sec:robust_linear_classification}
Assume the Polish space $S$ admits a decomposition $S=X\times Y$, where $X$ is a normed linear space: $(X, \norm{\marg})$ with $d_X(x,x') \defas \norm{x-x'}$ and $(Y,d_Y)$ is a metric space. We call $S$ the input--label space. We equip $S$ with the metric 
\begin{gather}
    d_S|_\kappa((x,y),(x',y')) \defas d_X(x,x') + \kappa d_Y(y,y'), \label{defn:kappa_metric}
\end{gather}
where $\kappa>0$.

\begin{toappendix}
    \autoref{lem:convex_concave} generalises \citet[Lemma~A.3]{shafieezadeh2017regularization} using a much simpler proof but essentially the same technique --- establishing the Lipschitzian constraint via Fenchel duality. \citet{sinha2017certifiable} use a similar condition, however it is assumed for technical convenience rather than derived analytically.
\end{toappendix}

\begin{appendixlemma}{lem:convex_concave}
    Let $(X,\norm{\marg})$ be a normed linear space with continuous dual $(X^*,\norm{\marg}^*)$ and $\Psi:X \to \Rx$. Then for all $z\in X$
    \begin{gather}
        \sup_{x \in X}\rbr\Big{\Psi(x) -\gamma\norm{x-z}} ≥ \Psi(z) + \infty·\iver*{\sup_{g^*\in\dom \Psi^*} \norm{g^*}^* > \gamma}.\label{eq:lip_ineq}
    \end{gather}
    If $\Psi$ is closed, proper and convex then we achieve equailty in \eqref{eq:lip_ineq}. If $\Psi$ is convex and Lipschitz then $\sup_{g^*\in\dom \Psi^*} \norm{g^*}^*= \lips(\Psi)$.
\end{appendixlemma}
\begin{proof}
    For every function $\Psi^{**}≤\Psi$ \citep[Proposition 3.43, p.~214]{penot2012calculus} and
    \begin{align}  %
        \sup_{x \in X}\rbr\Big{\Psi(x) -\gamma\norm{x-z}}
        &≥\sup_{x \in X}\rbr\Big{\Psi^{**}(x) -\gamma\norm{x-z}}\label{eq:cvxeq}
        \\&=\sup_{x \in X}\sup_{x^*\in\dom \Psi^*}\rbr\Big{\inp{x^*,x}-\Psi^{*}(x^*) -\gamma\norm{x-z}}
        \\&=\sup_{x \in X}\sup_{x^*\in\dom \Psi^*}\inf_{\norm{g^*}^*≤ \gamma}\rbr\Big{\inp{x^*,x}-\Psi^{*}(x^*) -\inp{g^*,x-z}}
        \\&=\sup_{x^*\in\dom \Psi^*}\sup_{x \in X}\inf_{\norm{g^*}^*≤
            \gamma}\rbr\Big{\inp{x^*-g^*,x}+
            \inp{g^*,z}-\Psi^{*}(x^*)}
        \\&= \sup_{x^*\in\dom \Psi^*}\inf_{\norm{g^*}^*≤ \gamma}\sup_{x \in X}\rbr\Big{\inp{x^*-g^*,x}+
    \inp{g^*,z}-\Psi^{*}(x^*)} \label{eq:penexc}
      \\&=\sup_{x^*\in\dom \Psi^*}\inf_{\norm{g^*}^*≤ \gamma}\rbr\Big{\infty·\iver{x^* ≠ g^*}+ \inp{g^*,z}-\Psi^{*}(x^*)}
        \\&=\sup_{x^*\in\dom \Psi^*}\rbr\Big{\infty·\iver*{\sup_{g^*\in\dom \Psi^*}\norm{g^*}^*> \gamma}+ \inp{x^*,z}-\Psi^{*}(x^*) }
        \\&=\Psi(z) + \infty·\iver*{\sup_{g^*\in\dom \Psi^*} \norm{g^*}^* > \gamma}.
    \end{align}
    The exchange of the supremum and infimum  in \eqref{eq:penexc} follows from \citet[Theorem~1.86, p.~59]{penot2012calculus}. 
    
    If $\Psi$ is closed, proper and convex then we achieve equality in \eqref{eq:cvxeq} \citep[Proposition 3.43, p.~214]{penot2012calculus}.
    
    The final claim is a standard result on convex functions \citep[e.g.][Lemma~2.6, p.~133]{shalev2012online}.
\end{proof}

\begin{theoremrep}\label{thm:multiclass_robust_dual_minimisation}
    Assume $d_S|_\kappa$ is finite, and lower semicontinuous with respect to the Polish topology on $S$.
    \begin{letblock}[before={Let }]
        \item $\ellf:S\to\R$ be continuous for all $f\in\cal F$
        \item $\ell_{f,y} \defas x\mapsto \ellf(x,y)$ is convex and Lipschitz for all $f\in\cal F$, $y\in Y$ 
        \item fix $\hat\mu(n)\in\probm(S)$ with $\hat\mu(n) \defas \frac1n \sum_{i=1}^n \dirac_{(x_i,y_i)}$
    \end{letblock}
    Then the robust risk minimisation
    \begin{gather}
        \minimise_{f\in\cal F} \risk(f;\costb_{d_S|_\kappa,\rho}(\hat\mu(n))),
    \intertext{has the same optimal value as the following problem:}
        \begin{rcases}\hspace{1em}
            \begin{alignedat}{3}
                \underset{\mathclap{\substack{f\in\cal F,\,\lambda\geq 0,\, l\in\R^n}}}{\mathrm{minimise}}
                &\hspace{3em}&&
                \frac1n\sum_{i=1}^n l_i +\lambda\rho
                \\
                \smash{\underset{\substack{\forall i = 1,\dots,n\\\forall y \in Y}}{\makebox[\widthof{$\mathrm{minimise}$}][c]{$\mathrm{subject\, to}$}}}&\hspace{3em}&&
                \ellf(x_i,y) - \lambda\kappa d_Y(y,y_i) ≤ l_i
                \\&&&\lips(\ell_{f,y_i})≤ \lambda.
            \end{alignedat}
        \end{rcases}\label{opt:robust_risk_dual}
    \end{gather}
\end{theoremrep}

\begin{proof}
    By \citet[Theorem~1]{blanchet2016quantifying} strong duality holds and
    \begin{align}
        \risk(f;\costb_{d_S|_\kappa,\rho}(\hat\mu(n)))
        = \inf_{\lambda\geq0}\rbr\Big{\lambda \rho + \int_S\hat\mu(n)(\d x)\sup_{s\in S}\rbr\Big{\ellf(s) - \lambda d_S|_\kappa(x, s)}}.
    \end{align}

    Using \autoref{lem:convex_concave} to evaluate the inner supremum we have
    \begin{align}
        \MoveEqLeft[8]\forall{i\in[n]}\sup_{s\in S}\rbr\Big{\ellf(s) - \lambda d_S|_\kappa\rbr{(x_i,y_i),s}}
        \\&= \sup_{(x',y')\in S} \rbr\Big{\ellf(x',y') - \lambda\norm{x' - x_i} - \lambda\kappa d_Y(y',y_i)}
        \\&= \sup_{y'\in Y}\sup_{x'\in X} \rbr\Big{\ellf(x',y') - \lambda\norm{x' - x_i} - \lambda\kappa d_Y(y',y_i)}
        \\&= \sup_{y'\in Y}\rbr\Big{\ellf(x_i,y') - \lambda\kappa d_Y(y',y_i) + \infty·\iver{\lips(\ell_{f,y_i})> \lambda}}.\label{eq:inner_sup_eval}
    \end{align}
    It follows that
    \begin{align}
        \MoveEqLeft[4]\inf_{\lambda\geq0}\rbr\Big{\lambda \rho + \frac1n\sum_{i=1}^n\sup_{s\in S}\rbr\Big{\ellf(s) - \lambda d_S|_\kappa\rbr{s_i,s}}}
        \\&\overset{\eqref{eq:inner_sup_eval}}{=}\mathrlap{\inf_{\lambda\geq0}\rbr\Big{\lambda \rho + \frac1n\sum_{i=1}^n\max_{y'\in Y}\rbr\Big{\ellf(x_i,y') + \infty·\iver{\lips(\ell_{f,y_i})>\lambda} - \lambda\kappa d_Y(y,y_i)}}}
        \\&\overset{\hphantom{\eqref{eq:inner_sup_eval}}}{=}\displaystyle\inf_{\lambda\geq0}\begin{cases}
            \lambda \rho + \displaystyle\frac1n\sum_{i=1}^n\max_{y'\in Y}\rbr\Big{\ellf(x_i,y') - \lambda\kappa d_Y(y,y_i)} & \displaystyle\max_{i\in [n]}\lips(\ell_{f,y_i})≤ \lambda\\
            \infty & \textrm{otherwise},
        \end{cases}
    \end{align}
    which yields
    \begin{align}
        \MoveEqLeft[4]\minimise_{f\in\cal F} \risk(f;\costb_{d_S,\rho}(\hat\mu(n)))
        \\&= \minimise_{f\in\cal F} \inf_{\lambda\geq0}\rbr\Big{\lambda \rho + \int_S\hat\mu(n)(\d x)\sup_{s\in S}\rbr\Big{\ellf(s) - \lambda c(x, s)}}
        \\&=\begin{cases}\hspace{1em}
            \begin{alignedat}{3}
                \underset{\mathclap{\substack{f\in\cal F,\,\lambda\geq 0,\, l\in\R^n}}}{\mathrm{minimise}}
                &\hspace{3em}&&
                \frac1n\sum_{i=1}^n l_i +\lambda\rho
                \\
                \smash{\underset{\substack{\forall i = 1,\dots,n\\\forall y \in Y}}{\makebox[\widthof{$\mathrm{minimise}$}][c]{$\mathrm{subject\, to}$}}}&\hspace{3em}&&
                \ellf(x_i,y) - \lambda\kappa d_Y(y,y_i) ≤ l_i
                \\&&&\lips(\ell_{f,y_i})≤ \lambda,
            \end{alignedat}
        \end{cases}
    \end{align}
    as claimed. 
\end{proof}

\begin{remark}\label{rem:no_label_noise_remark}
    For sufficiently large $\kappa$,  \eqref{opt:robust_risk_dual} becomes
    \begin{gather}
        \minimise_{f\in\cal F}    
        \frac1n\sum_{i=1}^n \ellf(x_i,y_i) + \rho \max_{i\in[n]}\lips(\ell_{f,y_i}).\label{eq:no_label_noise}
    \end{gather}
\end{remark}

\begin{example}[Multiclass logistic regression]\label{ex:softmax}
    The function $f_W:\R^n\to \probm([k])$ is a softmax layer parameterised by a linear operator $W:\R^n\to\R^k$. That is
    \begin{gather}
        f_W(x)\cbr{y} \defas \frac{\exp(Wx)_y}{\sum_{y'\in [k]}\exp(Wx)_{y'}},
    \end{gather}
    where the exponential function operates element-wise and the subscript indicates the $y$-th element. The loss function $\ell: \probm([k])\times [k]\to \R$ is the cross entropy error function (also called log loss):
        $\ell(\mu, y) \defas -\log\mu\cbr{y}$.
    Let $\ell_{f_W}: \R^n\times [k]\to\R$ with $\ell_{f_W}(x,y) \defas \ell(f_W(x), y)$. Then
    \begin{align}
        \ell_{f_W}(x,y) = -(Wx)_y  + \log\sum_{y'\in Y}\exp(Wx)_{y'}
    \end{align}
    is convex in $x$ and $\lips_{\R^n}(\ell_{f_W}) ≤ \opnorm{W}$, the operator norm of $W$. 
\end{example}

\section{Robust deep classification}\label{sec:robust_deep_classification}

\begin{toappendix}
    \begin{figure}
        \centering
        \begin{tikzpicture}
            \tikzstyle{every node}=[font=\small]
            \coordinate (muhat) at (-2.5,0);
            \coordinate (phimuhat) at (2.5,0);
    
            \draw [red, fill = red!10!white] (muhat) circle[radius = 3em] ;
            \draw (muhat) edge[red, <->, shorten <= 2pt] node[midway, above, sloped] {\tiny$\rho$} + (225:3em);
            \draw (muhat)node[dot] {} node[anchor = south] {$\mu$};
            
            \coordinate (pimuhatellipse) at ($(phimuhat) + (0.25,0.25)$);
            \draw[blue, fill = blue!10!white] (phimuhat) circle (3.75em);
            \draw[red, fill = red!10!white] (pimuhatellipse) ellipse (3em and 2em);
            \draw (phimuhat) node[dot] {} node[anchor = south, yshift=3pt] {$\phi_\#\mu$};
    
            \draw (muhat) edge[black, dashed,-latex, bend left=15, shorten >= 2pt] node[midway, anchor = south] {$\phi_\#$} (phimuhat) ;
    
            \draw (phimuhat) edge[blue, <->, shorten <= 2pt] node[midway, above, sloped] {\tiny$\rho\lips(\phi)$} + (-45:3.8em);
    
            \draw[dashed, red] (muhat) ++ ( 90:3em) edge ($(pimuhatellipse) + ( 90:3em and 2em)$);
            \draw[dashed, red] (muhat) ++ (-90:3em) edge ($(pimuhatellipse) + (-90:3em and 2em)$);
    
            \draw[red] (muhat)     ++ (90:3em) node[anchor = south] {$\costb_{d_S,\rho}(\mu)$}; 
            \draw[blue] (phimuhat) ++ (90:3.75em) ++ (0.1, 0) node[anchor = south] {$\costb_{d_T,\rho\lips(\phi)}(\phi_\#\mu)$}; 
    
            \draw[gray,thick, dotted] (muhat)    ++ (-1,0) ++ (40:3) arc (40:-40:3);
            \draw[gray,thick, dotted] (phimuhat) ++ (1,0)  ++ (220:3) arc (220:140:3);
            \draw[gray] (muhat)    ++ ( 0.8,-1.7) node {\large$\probm(S)$};
            \draw[gray] (phimuhat) ++ (-0.8,-1.7)  node {\large$\probm(T)$};
        \end{tikzpicture}
        \caption{Illustration of \autoref{lem:wasserstein_pushforward}.\label{fig:wasserstein_pushforward}}
    \end{figure}        
\end{toappendix}

In order to extend the results from \autoref{sec:robust_linear_classification} to a more rich class of models, namely deep neural networks, we lift the input space $X\times Y$ via a nonlinear Lipschitz transformation $\phi$ to a feature space $Z\times Y$. In \autoref{lem:wasserstein_pushforward} we push a neighbourhood of the original distribution $\mu\in\probm(X\times Y)$ to a neighbourhood of $\phi_\#\mu\in\probm(Z\times Y)$. This is illustrated in \autoref{fig:wasserstein_pushforward}. In the feature space, the assumptions needed  for \autoref{thm:multiclass_robust_dual_minimisation} are compatible with modern deep architectures and are no longer onerous.

\begin{appendixlemma}{lem:wasserstein_pushforward}
    \begin{letblock}
        \item $(S,d_S)$ and $(T,d_T)$ be metric spaces
        \item $\phi\in\bmeas(S;T)$ a Lipschitz mapping
    \end{letblock}
    Then for $\mu\in\probm(S)$
    \begin{gather}
        \cbr{\phi_\#\nu\given\nu\in\costb_{d_S,\rho}(\mu)}\subseteq\costb_{d_T,\rho\lips(\phi)}(\phi_\#\mu).
    \end{gather}
\end{appendixlemma}

\begin{proof}
    Let $\nu\in\costb_{d_S,r}(\mu)$. Then from the Lipschitz continuity of $\phi$ we have
    \begin{align}
        \lips(\phi)\rho
        &\geq\lips(\phi)\cost_{d_S}(\mu,\nu) 
        \\&= \inf_{\pi\in\coup\rbr{\mu,\nu}}\int_S \lips(\phi) d_S(s, s')\d \pi
        \\&\geq \inf_{\pi\in\coup\rbr{\mu,\nu}}\int_T d_T(\phi(s),\phi(s'))\d \pi
        \\&= \inf_{\pi\in\coup\rbr{\mu,\nu}}\int_T d_T(z,z')\d (\phi,\phi)_\#\pi
        \\&\geq \inf_{\pi\in\coup\rbr{\phi_\#\mu,\phi_\#\nu}}\int_T d_T(z,z')\d \pi,
    \end{align}
    where the last inequality follows since $\cbr{(\phi,\phi)_\#\pi : \pi\in\coup\rbr{\mu,\nu}}\subseteq \coup\rbr{\phi_\#\mu,\phi_\#\nu}$.
    This shows that $\phi_\#\nu \in \costb_{d_T,\rho\lips(\phi)}(\phi_\#\mu)$ and completes the proof.
\end{proof}


Define
\begin{letblock}[before = {}]
    \item the input (metric) space $(X, d_X)$
    \item label (metric) space, $(Y, d_Y)$
    \item the feature (normed, linear) space $(Z, \norm{\marg})$ and continuous dual $(Z^*, \norm{\marg}^*)$,  with $d_Z(z,z') \defas \norm{z-z'}$
\end{letblock} Equip the input--label space $S\defas X\times Y$ with the metric 
\begin{gather}
    d_S|_\kappa((x,y),(x',y')) \defas d_X(x,x') + \kappa d_Y(y,y'). \tag{\ref{defn:kappa_metric}}\label{eq:kappa_metric_2}
\end{gather}\noeqref{eq:kappa_metric_2}
The metric on the feature--label space $T\defas Z\times Y$ is 
\begin{gather}
    d_T|_\gamma((z, y), (z',y')) \defas d_Z(z,z') + \gamma d_Y(y,y'),
\end{gather}
which we assume finite. These spaces and the notation for the functions between them are summarised in \autoref{fig:commutativity}.

\begin{figure}
    \centering
    \begin{tikzpicture}
        \node (x) at (0,0) {$(X,d_X)$} node[below = 1em of x] (y) {$(Y,d_Y)$};
        \path (x) -- (y)  node[midway] {$\times$};

        \node[right = 5em of x] (z) at (0,0) {$(Z,\norm{\marg})$} node[below = 1em of z] (y2) {$(Y,d_Y)$};
        \node[right = of z] (v) {$V$};
        \path (z) -- (y2)  node[midway] (m) {$\times$};

        \draw[rounded corners, dashed] (x.north west) rectangle (y.south east) node[above = 0.25em of x] (s) {$(S,d_S|_\kappa)$};
        \draw[rounded corners, dashed] (z.north west) rectangle (y2.south east) node[above = 0.25em of z] (t) {$(T,d_t|_\gamma)$};

        \path (x) edge[->, shorten <= 3pt, shorten >= 3pt] node[midway,above] {$\vphi$} (z);
        \path (z) edge[->, shorten <= 3pt, shorten >= 3pt] node[midway,above] {$h$} (v)    ; 
        \path (y) edge[->, shorten <= 3pt, shorten >= 3pt] node[midway,above] {$\id$}  (y2); 
    \end{tikzpicture}
    \caption{Commutativity diagram for the spaces and functions identified in \autoref{sec:robust_deep_classification}.\label{fig:commutativity}}
\end{figure}

\begin{theoremrep}\label{thm:pushforward_risk}
    \begin{letblock}
        \item $(S, d_S|_\kappa)$ and $(T, d_T|_\gamma)$ be as defined above
        \item $\vphi\in\bmeas(X;Z)$ is Lipschitz
        \item $\ell_{h}\in\bmeas(Z;\Rx)$
        \item assume $\ell_{h\circ\vphi} = \ell_{h}\circ(\vphi,\id)$
    \end{letblock}
    Then $(\vphi,\id):S\to T$ is Lipschitz, and for all $\mu\in\probm(S)$
    \begin{align}
        \risk(h\circ\vphi;\costb_{d_S|_\kappa,\rho}(\mu))≤ \risk(h;\costb_{d_T|_{\kappa\lips(\vphi)},\rho\lips(\vphi)}((\vphi,\id)_\#\mu)).\label{defn:push_forward_risk}
    \end{align}
\end{theoremrep}
\begin{proof}
    The only subtlety is in showing that the mapping $(\vphi,\id):S\to T$ is Lipschitz with constant $\lip(\vphi)$. Since $\vphi$ is Lipschitz we have
    \begin{gather}
        \forall{x,x'\in X}d_Z(\vphi(x),\vphi(x'))≤ \lips(\vphi)d_X(x,x'),
        \shortintertext{and thus for all $x,x'\in X$ and all $y,y'\in Y$} 
        \begin{aligned}
           d_Z(\vphi(x),\vphi(x')) + \kappa\lips(\vphi) d_Y(y,y')
            &≤ \lips(\vphi)d_X(x,x') +  \kappa\lips(\vphi)d_Y(y,y')
            \\&= \lips(\vphi)\rbr\big{d_X(x,x') +\kappa d_Y(y,y')}.
        \end{aligned}
        \shortintertext{This is equivalent to}
        \forall{s,s'\in S} d_T|_{\gamma}((\vphi,\id)(s),(\vphi,\id)(s')) ≤ \lips(\vphi)d_S|_\kappa(s,s'),
    \end{gather}
    where $\gamma\defas\kappa·\lips(\vphi)$. Thus $(\vphi,\id)$ is $\lips(\vphi)$-Lipschitz with respect to $d_T|_\gamma$. The theorem then follows from \autoref{lem:wasserstein_pushforward}.
\end{proof}
In light of \autoref{thm:pushforward_risk}, we call the right hand side of \eqref{defn:push_forward_risk} the \emph{pushforward (robust) risk}. Under a suitable convexity assumption on $\ell_{h}$ we can minimise the pushforward risk using \autoref{thm:multiclass_robust_dual_minimisation}.

\begin{propositionrep}\label{prop:deep_robust_classifcation}
    Let $(S, d_S|_\kappa)$ and $(T, d_T|_\gamma)$ be as defined above, $V$ is an appropriate prediction (metric) space.
    Assume $T$ is a Polish space and $d_T|_\gamma$ is finite, and lower semicontinuous with respect to the Polish topology.
    \begin{letblock}[before={Assume }]
        \item $\cal F$ is a collection of mappings $f\in\bmeas(X;V)$ that admit a decomposition $f= h\circ\vphi$, where $\vphi\in\bmeas(X;Z)$ and $h\in \bmeas(Z;V)$
        \item $\ell_{h}\in\bmeas(T;\R)$ with $\ell_{h,y}(z) \defas \ell_h(z,y)$ is convex and Lipschitz for all $\cbr{h:h\circ\vphi \in \cal F}$, $y\in Y$ 
    \end{letblock}
    Fix $\hat\mu(n)\in\probm(S)$ with $\hat\mu(n) \defas \frac1n \sum_{i=1}^n \dirac_{(x_i,y_i)}$.
    The pushforward risk minimisation problem
    \begin{gather}
       \minimise_{h\circ\vphi\in\cal F} 
       \risk(h;\costb_{d_T|_{\kappa·\lips(\vphi)},\rho\lips(\vphi)}((\vphi,\id)_\#\hat\mu(n))),
    \intertext{has the same optimal value as the following problem:}
        \begin{rcases}\hspace{1em}
            \begin{alignedat}{3}
                \underset{\mathclap{\substack{h\circ\vphi\in\cal F,\,\lambda\geq 0,\, s\in\R^n}}}{\mathrm{minimise}}
                &\hspace{3em}&&
                \frac1n\sum_{i=1}^n s_i +\lambda\rho
                \\
                \smash{\underset{\substack{\forall i = 1,\dots,n\\\forall y \in Y}}{\makebox[\widthof{$\mathrm{minimise}$}][c]{$\mathrm{subject\, to}$}}}&\hspace{3em}&&
                \textstyle\ell_{h\circ\vphi}(x_i,y) - \lambda\kappa\lips(\vphi)d_Y(y,y_i) ≤ s_i
                \\&&&\textstyle\lips(\ell_{h,y_i})\lips(\vphi)≤ \lambda.
            \end{alignedat}
        \end{rcases}
    \end{gather}
\end{propositionrep}

From \autoref{ex:softmax} it's clear that cross-entropy loss composed with a softmax layer satisfies the conditions in \autoref{prop:deep_robust_classifcation} on $\ell_h$ (convex, Lipschitz), and the composite mapping of the hidden layers satisfies the conditions on $\vphi$ (Lipschitz).

\subsection{Lipschitz networks}\label{ssec:lipschitz_networks}

Let $(X_i, d_{i})_{i=1}^{l+1}$ be a sequence of metric spaces with the special identifications of the input space, $X \defas X_1$, and the prediction space, $V \defas X_{l+1}$. We write a neural network $\vphi:X\to V$ using a sequence of mappings $\vphi_{i}:X_i\to X_{i+1}$ with $\vphi \defas \vphi_1\circ \dots \circ \vphi_{l}$. Let $\neuralnet^l(X;V)\subseteq \bmeas(X;V)$ be the collection of $l$-layer composite mappings that satisfy $\lips(\vphi)<\infty$ for all $\vphi \in \neuralnet^l(X;V)$. The network $\vphi$ is learnt by minimising the risk \eqref{defn:distributionally_robust_risk}  defined with a loss function $\ell:V\times Y\to\Rx$. 

Fix $\vphi\in\neuralnet^l(X;V)$. The minimal Lipschitz constant satisfies $\lips(\vphi) ≤ \prod_{i=1}^l \lips(\vphi_i)$. Each function $\vphi_i$ may be further broken down into the composition of a linear operator $W_i:X_i\to X_{i}'$ and a nonlinear activation function $\alpha_i:X_{i}'\to X_{i+1}$ with $\vphi_i \defas \alpha_i \circ W_i$ and $\lips(\vphi_i) ≤ \lips(\alpha_i)\lips(W_i)$. For linear operators between normed spaces (that is $d_i(x,y) = \norm{x-y}_i$ for some norm $\norm{\marg}_i$)  we have the convenient representation in terms of the operator norm $\lip(W_i) = \opnorm{W_i}$ where $ \opnorm{W_i} \defas \sup_{\norm{x}_{i}≠1} \norm{W_i x}_{i+1}$. Since Lipschitz constants for typically used activation functions are less than 1 \citep{yoshida2017spectral,miyato2018spectral,gouk2018regularisation}, using the Young product inequality we have
\begin{gather}
    \lips(\vphi) ≤ \prod_{i=1}^l\lips(\vphi_i) ≤ \prod_{i=1}^l \lips\rbr{\alpha_i}·\prod_{i=1}^l \opnorm{W_i} ≤ \frac1l\sum_{i=1}^l \opnorm{W_i}^l.
    \label{eq:lipschitz_constant_of_neural_network}
\end{gather}

\begin{remark}
    The remark analogous to \autoref{rem:no_label_noise_remark} for \autoref{prop:deep_robust_classifcation} with a deep neural network is
    \begin{gather}
        \minimise_{\substack{\vphi_{1:l-1} \in \neuralnet^{l-1}(X;Z),\\\vphi_l \in \neuralnet^1(Z;V)}}
        \frac1n\sum_{i=1}^n \ell_\vphi(x_i,y_i) + \rho\max_{i\in[n]}\lips(\ell_{\vphi_l,y_i})\lips(\vphi_{1:l-1}).\label{eq:input_robust_risk}
    \end{gather}
    Since Lipschitz constant can be upper bounded by the operator norms of the weight matrices, using \eqref{eq:lipschitz_constant_of_neural_network} we obtain the upper bound on \eqref{eq:input_robust_risk}:
    \begin{gather}
        \frac1n\sum_{i=1}^n \ell_\vphi(x_i,y_i) + \rho\lips_X(\ell) \prod_{j=1}^l \lips\rbr{\vphi_j} ≤\frac1n\sum_{i=1}^n \ell_\vphi(x_i,y_i) + \frac{\rho\lips_X(\ell)}{l}\sum_{j=1}^l\opnorm{W_j}^{l}.\label{eq:spec_reg}
    \end{gather}
\end{remark}

The tightness of the left-hand bound \eqref{eq:spec_reg} over \eqref{eq:lipschitz_constant_of_neural_network} is undoubtably related to the number of layers $l$ for with every composition step we introduce the possibility of slackness: $\lips(\vphi_j\circ \vphi_{j-1})\leq\lips(\vphi_j)\lips(\vphi_{j-1})$. A natural solution to this problem for deep neural networks is to constrain the individual layer Lipschitz constants to be all less than 1 so that $ \prod_{j=1}^l \lips\rbr{\vphi_j}$ does not explode \citep[cf.][]{cisse2017parseval}. The right-hand bound in \eqref{eq:spec_reg} is virtually identical to the spectral regularisation proposed by \citet[viz.\ equation 1]{yoshida2017spectral}, but with a complete intuition for the their regularisation parameter $\lambda$ and a distributional robustness guarantee via \autoref{thm:pushforward_risk}. The particular case of the spectral norm corresponds to the $\leb_2$ norm on $X$, $Z$, and $V$.

\section{Adversarial robustness}\label{ssec:distributional_robustness_is_adversarial_robustness}

The following objective function has been proposed \citep{goodfellow2015explaining,madry2017towards,shaham2018understanding} in order to build a robust classifier $f$ on an input--label space $X\times Y$:
\begin{gather}
    \int_{X\times Y} \mu(\d x, \d y) \max_{\epsilon \in B} \ellf(x + \epsilon, y),
    \label{eq:madryetal}
\end{gather}
where the set $B\subseteq X$ is typically an $\leb_p$ ball. We refer to \eqref{eq:madryetal} as the \emph{adversarial risk}. With \autoref{thm:adversarial_risk_bound} we see that distributionally robust risk minimisation also minimises the adversarial risk.
\begin{appendixlemma}{lem:adversarial_risk_bound}
    \begin{letblock}[before={Fix }]
        \item a topological space $S$
        \item probability measure $\mu\in\probm(S)$
        \item cost function $c\in\bmeas(S\times S;\Rx)$
    \end{letblock}
    If $a\in\bmeas(S;S)$ satisfies $\E_\mu c\circ (\id\times a) ≤  \rho$, then $ a_\#\mu \in \costb_{c,\rho}(\mu)$.
\end{appendixlemma}
\begin{proof}
    Since $(\id\times a)_\#\mu\in\coup(\mu,a_\#\mu)$ we have
    \begin{align}
        \cost_{c}(\mu,a_\#\mu) 
        = \inf_{\pi\in\coup(\mu,a_\#\mu)} \int c \d \pi
        ≤ \int c \d (\id \times a)_\#\mu
        = \int c(x,a(x)) \mu(\d x)
        ≤ \rho,
    \end{align}
    which shows $ a_\#\mu \in \costb_{c,\rho}(\mu)$.
\end{proof}
\begin{theoremrep}\label{thm:adversarial_risk_bound}
    \begin{letblock}[before={Fix }]
        \item a separable Banach space $S$
        \item cost function $c(x,y)=\norm{x-y}$ for some $\norm{\marg}\in\bmeas(S;\Rx)$
        \item probability measure $\mu\in\probm(S)$
        \item $\ellf:S\to\Rx$ is upper semicontinuous
        \item $A\subseteq S$ is closed
    \end{letblock}
    Then
    \begin{gather}
        \int_S \mu(\d s) \sup_{a \in A} \ellf(s + a)≤ \risk(f; \costb_{c,\rho}(\mu)),\label{eq:ball_risk_bound}
    \end{gather}
    where $\rho \defas \sup_{a\in A}\norm{a}$.
\end{theoremrep}
\begin{proof}
    Let $\Gamma(s) \defas A + s$. Since $A$ is closed, $\Gamma$ is closed and Borel measurable. Let $\bmeas_\mu(\Gamma) \defas \cbr{g\in\bmeas(S;S) :g(s) \in \Gamma(s)\; \text{$\mu$-a.e.}}$.  The set $\bmeas_\mu(\Gamma)$ is trivially decomposable \citep[Definition 2.1]{giner2009necessary}. By assumption $\ellf$ is upper semicontinuous and therefore Borel measurable. Since $\ellf$ is measurable, any decomposable subset of $\bmeas(S;S)$ is $\ellf$-decomposable \citep[Proposition 5.3]{giner2009necessary} and $\ellf$-linked \citep[Proposition 3.7 (i)]{giner2009necessary}.
    Using Theorem 6.1 (c) of \citet{giner2009necessary} we have
    \begin{align}
        \int \mu(\d s) \sup_{a \in A} \ellf(s + a)
        =\int_S \mu(\d s) \sup_{s'\in\Gamma(s)}\ellf(s') 
        = \sup_{x\in \bmeas_\mu(\Gamma)}\int_S\mu(\d s)(\ellf\circ x)(s).\label{eq:giner}
    \end{align}

    Let $g\in \bmeas_\mu(\Gamma)$. Then for $\mu$-almost all $s\in S$
    \begin{gather}
        c(s, g(s))  = \norm{s- (h+\id)(s)}= \norm{s - s + h(s)} =\norm{h(s)} ≤  \sup_{a\in A}\norm{a} = \rho. 
    \end{gather}
    We then apply \autoref{lem:adversarial_risk_bound} to $g$ obtain the bound
    \begin{gather}
        \mathclap{\int \mu(\d s) \sup_{a \in A} \ellf(s + a) \overset{\eqref{eq:giner}}{=} \sup_{g\in \bmeas_\mu(\Gamma)}\int_S  g_\#\mu(\d s) \ellf(s) 
        ≤ \sup_{\nu\in\costb_{c,\rho}(\mu)}\int_S  \nu(\d s) \ellf(s).}\label{eq:forwards_inequality}
    \end{gather}
    This completes the proof of \eqref{eq:ball_risk_bound}.
    %
    %
    %
\end{proof}

\begin{remark}
    In at least some cases it is possible to strengthen \eqref{eq:ball_risk_bound} to an equality. \citet[Theorem 3.20]{shafieezadeh2017regularization} achieve equality in \eqref{eq:ball_risk_bound} for a realization $\hat\mu(n)$ in the setting $S \defas \R^n\times\cbr{-1,+1}$, linear classifiers $f$, $\ellf(x,y) = \ell(y·f(x))$ for some Lipschitz  $\ell:\R\to\R$, and where the cost function takes on the value of $\infty$ in the case of label noise, that is, $\kappa=\infty$ in \eqref{defn:kappa_metric}. 
\end{remark}

\begin{remark}
    Let $S\defas X\times Y\defas \R^{n}\times([k]\cup\cbr{0})$. Let $A\times \cbr{0} \subseteq X \times Y$. If we use the $\kappa$-metric \eqref{defn:kappa_metric} for the cost function with $d_X(x,x')=\norm{x-x'}$ then for all $\kappa>0$ we obtain exactly \eqref{eq:madryetal} in the left hand side of \eqref{eq:ball_risk_bound}. Furthermore with a sufficiently large $\kappa$ we obtain the objective function \eqref{eq:input_robust_risk}. This is very similar to the Lipschitz bound on the adversarial risk noted by  \citet[\S3.2]{cisse2017parseval}.
\end{remark}

\section{Discussion}\label{sec:discussion}

The idea that there should be a relationship between the Lipschitz continuity of a neural network and its robustness to adversarial attack is not new. \citeauthor{szegedy2013intriguing}, in their landmark paper on adversarial learning \citeyearpar{szegedy2013intriguing}, conclude with a section where they study the Lipschitz properties of ImageNet. Other papers have since attempted to formalise the connection between Lipschitz continuity and adversarial robustness: 
\citet{sinha2017certifiable} --- like us --- work with the dual problem of \eqref{defn:distributionally_robust_risk}, however the distributional certificate they compute is difficult to work with in practice and as a result \citeauthor{sinha2017certifiable} resort to heuristic choices for hyperparameters; \citet{tsuzuku2018lipschitz} use the Lipschitz constant of the network to provide a margin-based, local certification of adversarial robustness (Propositions 1 and 2) for multiclass classification; \citet{huster2018limitations} prove the existence of a Lipschitz, adversarially-robust binary classifier; and \citet{suggala2018adversarial} bound the adversarial risk (binary classification) in the special case of the logistic loss using a term similar to Lipschitz constant (Theorem 7). Two more recent attempts at robustness certificates can be found in the papers of \citet{raghunathan2018certified} and \citet{wong2018provable}, both of which are neither closed form nor easy to compute. 

\citet{cisse2017parseval} comes the closest to our result, bounding the adversarial risk using the Lipschitz constant of the network. However, the certificate in \autoref{sec:robust_deep_classification} bounds the much more general concept of distributionally robust risk, rather than the special (albeit pathological) case of the adversarial risk. Additionally, the Lipschitz constant can also be related to model complexity \citep{bartlett2017spectrally} which lends a natural interpretation of the objective function in \eqref{eq:input_robust_risk} with respect to a trade-off of goodness of fit and parsimony.

\section{Conclusion}\label{sec:conclusion}
Lipschitz-constrained networks have been shown to train faster, and have better robustness and generalisation properties \citep{cisse2017parseval,yoshida2017spectral,miyato2018spectral,gouk2018regularisation}. We have directly shown the relationship between Lipschitz continuity and distributional robustness to explain the success of theses approaches. In application we hae seen that adversarial robustness is implicitly a special case of distributional robustness. These results give the theoretical basis of Lipschitz regularisation to build better, more robust models.

\section*{Acknowledgements}

The authors wish to thank Olivier Bousquet, Ian Goodfellow, and Erik Davis for their helpful comments. 

\bibliography{bibliography}

\appendix

\end{document}